\documentclass[10pt,twocolumn,letterpaper]{article}

\usepackage{iccv}
\usepackage{times}
\usepackage{epsfig}
\usepackage{graphicx}
\usepackage{amsmath}
\usepackage{amssymb}
\usepackage{ulem}
\usepackage{mathrsfs}
\usepackage[table,xcdraw]{xcolor}
\usepackage{color}
\usepackage{url}
\usepackage{siunitx}
\usepackage{booktabs}
\usepackage{tabularx}
\usepackage{float}
\usepackage{algorithm}
\usepackage{algpseudocode}
\usepackage{amsmath}
\usepackage{multirow}

\usepackage[pagebackref=true,breaklinks=true,letterpaper=true,colorlinks,bookmarks=false]{hyperref}

\def\iccvPaperID{6225} 

\iccvfinalcopy

\begin{document}

\title{Free-form Description Guided 3D Visual Graph Network for Object Grounding 
in Point Cloud}

\author{Mingtao Feng$^{1,2}$\thanks{Equal contribution}~~ Zhen Li$^{1 *}$~~ Qi Li$^{1 *}$~~ Liang Zhang$^{1 }$\thanks{Corresponding author}~~ XiangDong Zhang$^{1}$~~ Guangming Zhu$^{1}$ \\Hui Zhang$^{2}$~~ Yaonan Wang$^{2}$~~ Ajmal Mian$^{3}$\\
	$^1$Xidian University,~~ $^2$Hunan University,~~ $^3$The University of Western Australia\\
}

\maketitle

\ificcvfinal\thispagestyle{empty}\fi

\begin{abstract}
\vspace{-4mm}
3D object grounding aims to locate the most relevant target object in a raw point cloud scene based on a free-form language description. Understanding complex and diverse descriptions, and lifting them directly to a point cloud is a new and challenging topic due to the irregular and sparse nature of point clouds. There are three main challenges in 3D object grounding: 
to find the main focus in the complex and diverse description; 
to understand the point cloud scene; and 
to locate the target object. 
In this paper, we address all three challenges.
Firstly, we propose a language scene graph module to capture the rich structure and long-distance phrase correlations.
Secondly, we introduce a multi-level 3D proposal relation graph module to extract the object-object and object-scene co-occurrence relationships, and strengthen the visual features of the initial proposals. 
Lastly, we develop a description guided 3D visual graph module to encode global contexts of phrases and proposals by a nodes matching strategy. 
Extensive experiments on challenging benchmark datasets (ScanRefer~\cite{3DVL_Scanrefer} and Nr3D~\cite{InstanceRefer}) show that our algorithm outperforms existing state-of-the-art. Our code is available at \url{https://github.com/PNXD/FFL-3DOG}. 
\end{abstract}

\vspace{-5mm}
\section{Introduction}
\vspace{-2mm}
Imagine a scenario where an old person with limited mobility wakes up in the morning, feeling unwell, and instructs a robot to fetch medicine from a brown table. He/she could say ``\textit{A brown table is located in the corner of the room, it is to the right of a white cabinet and to the left of black shoes. The front of it is a light blue curtain.}'' 
For a human, finding the table based on the free-form language is an easy task. However, for assistive robots, parsing the large 3D visual scene, finding the target object and understanding the global context based on natural language descriptions is a challenging task. These sentences describe the appearance of the target object (\textit{table}), its spatial location relative to other objects (\textit{cabinet}, \textit{curtain} and \textit{shoes}) and the global scene (\textit{room}), which offer a rich source of information to localize the target object and guide the robot.

\begin{figure}[] 
	\center{\includegraphics[width=0.45\textwidth]{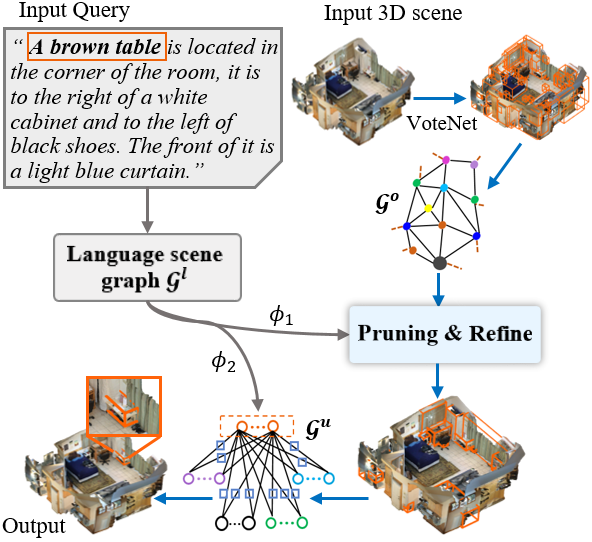}}
	\caption{\label{fig:Sumary} Proposed model for object grounding in 3D scenes. A multi-level proposal relation graph $\mathcal{G}^o$ is formed to strengthen the visual features of the initial proposals, then the 3D visual graph $\mathcal{G}^u$ is constructed under the guidance of the language scene graph $\mathcal{G}^l$ which refines the initial coarse proposals. The language scene graph $\mathcal{G}^l$ predicts the nodes matching with the 3D visual graph $\mathcal{G}^u$ and the matching scores $\phi_1$ and $\phi_2$ are fused to make the final grounding predictions.}
	\vspace{-4mm}	
\end{figure}

With the widespread availability of LiDARs, depth cameras and light field cameras~\cite{Survey_guo,RGN_3D}, 3D scene representation in the from of point clouds is becoming increasingly available and affordable in many application domains such as robotics, autonomous driving etc. Understanding the free-form descriptions and lifting them to the 3D point cloud scene is a new topic and challenging in the field of vision-and-language.

Research on the use of complex descriptions for 3D object grounding in point clouds is still in its infancy and only a few methods exist in the literature. To advance this line of research, ScanRefer~\cite{3DVL_Scanrefer} introduced the first large-scale dataset that couples free-form descriptions to objects in 3D scenes. The ScanRefer~\cite{3DVL_Scanrefer} method consists of two stages: the first stage aims to use a 3D object detector VoteNet~\cite{VoteNet} to generate a set of 3D object proposals based on the input point cloud, the second stage correlates a global language expression with 3D visual features of each proposal, computes a confidence score for each fused feature, and then takes the proposal with the highest score as the target object. Similar to ScanRefer~\cite{3DVL_Scanrefer}, Yuan et al.~\cite{InstanceRefer} replaced the 3D object detector with 3D panoptic segmentation backbone to obtain instance-level candidates, and captured the context of each candidate for visual and language feature matching, where the instance with the highest score is regarded as the target object.

The above methods suffer from several issues due to the inherent difficulties of both complex language processing and irregular 3D point cloud recognition. Firstly, free-form descriptions are complex and diverse in nature, and contain several sentences where strong context and long-term relation dependencies exist among them. Both ScanRefer~\cite{3DVL_Scanrefer} and InstanceRefer~\cite{InstanceRefer} do not consider that and only learn holistic representations of the description. Secondly, these two-stage solutions heavily rely on 3D object detectors or 3D panoptic segmentation backbones, and the quality of the object proposals obtained in the first stage are coarse which severely affect the overall performance. Conventional two-stage methods use the coarse object proposals directly in the following steps and fail to consider the relationships between the surrounding object proposals and global information in cluttered 3D scenes to refine the visual features of coarse proposals. Thirdly, the relationships between object proposals and phrases have not been fully explored. All existing methods~\cite{3DVL_Scanrefer,InstanceRefer} neglect linguistic and visual structures, and only fuse the global input description embedding with 3D visual features for grounding.  

To address the aforementioned limitations, we propose a free-form description guided 3D visual graph network for 3D object grounding in point clouds, as shown in Figure~\ref{fig:Sumary}. Especially, to incorporate the rich structure and language context, we parse the complex free-form description into three components (noun phrases, pronouns and relation phrases) and construct a language scene graph to compute context-aware phrase presentation through message propagation, in which the nodes and edges correspond to noun phrase plus pronouns and relation phrases respectively. Moreover, through the 3D object detector of VoteNet~\cite{VoteNet}, a set of initial 3D object proposals are extracted from the input raw point cloud.
We introduce a multi-level 3D relation graph to leverage two co-occurrence relationships (object-object and object-scene), which strengthen the visual features of the initial proposals to boost the performance of the subsequent operations. 
Furthermore, we use the language scene graph to guide the pruning of redundant proposals and then refine the selected ones. Built on top of the refined proposal set, we introduce a 3D visual graph to generate a context-aware object representation via message propagation. Concretely, nodes are the selected proposals relevant to the noun phrase, and the edges encode the relationships between object proposals. Finally, the nodes of 3D visual graph are adaptively matched with the nodes of language scene graph, then fused with the matching score in proposals pruning for the final 3D object grounding.

To sum up, our key contributions include: (1) We propose a free-form description guided 3D visual graph for object grounding that directly exploits the raw point cloud and is end-to-end trainable. 
(2) We propose a language scene graph module that captures the rich structure and long-distance phrase correlations;
(3) We propose a multi-level 3D proposal relation graph module that extracts the object-object and object-scene co-occurrence relationships to strengthen the visual features of the initial proposals; 
(4) We propose a description 3D visual graph module that encodes global contexts of phrases and proposals through nodes matching. 
Experiments were performed on the benchmark ScanRefer~\cite{3DVL_Scanrefer} and Nr3D~\cite{3DVL_Scanrefer} datasets and state-of-the-art results~\cite{3DVL_Scanrefer,ReferIt3D,InstanceRefer} were achieved.


\vspace{-2mm}
\section{Related work}
\vspace{-2mm}
\noindent {\bf 2D object grounding in images:} 
2D visual grounding aims to localize objects in an image corresponding to noun phrases from a given language description~\cite{GD-review,Cops-Ref-data}.
A significant number of 2D visual grounding works can be found in the literature that focus on bounding-box-level~\cite{DynamicGA,Neighbor_Watch,RealT-GD,Model_context,Sequence-GD,Corss-modal} and pixel-level~\cite{Mattnet,Seg-GD1,Seg-GD2} comprehensions from the input language description. 
Most 2D object grounding approaches follow a two stage approach where an pre-trained object detector like Faster RCNN~\cite{FasterRCNN} is first used to generate a set of 2D bounding box candidates based on the input image, and then the referred object is predicted in the second stage depending on the ranked matching score between each 2D bounding box candidate and the query sentence. 
However, most 2D object grounding methods mainly focus language description comprising a single sentence~\cite{Referitgame,Flickr30k,Generation}. Although impressive progress has been made in the field of 2D object grounding, there is little work on lifting natural language processing tasks to 3D point clouds. Effective modeling of 3D vision-language tasks require establishing carefully designed connections between language and the 3D point cloud data. Whereas 2D visual grounding methods can provide some guidance for 3D relationships learning, they can not be directly used for 3D object grounding in point clouds. 

\vspace{1mm}
\noindent {\bf 3D Object Detection in Point Clouds:} 
Point clouds are usually converted to canonical forms such as 2D images~\cite{Multi-view,Single-view,3Dobject-survey} or regular grids~\cite{Voxelnet,Pixor,Pointpillars} for 3D object detection with Convolutional Neural Networks. Recently, Qi et al.~\cite{VoteNet} proposed a framework to process raw point clouds directly and then predict 3D bounding boxes in cluttered scenes via a combination of PointNet++~\cite{PointNet++} backbone and Hough voting. However, VoteNet~\cite{VoteNet} focuses on regressing each 3D object independently and the local geometric information is not well accounted for. Chen et al.~\cite{Hierarchical} proposed a hierarchical graph network to aggregate features and capture shape information of objects in point clouds. Point-GNN~\cite{PointCNN} differs from previous works by introducing a graph neural network to detect the category and shape of 3D objects that each nodes in the graph belongs to. Since the 3D real world scene contains partially scanned objects with physical connections, dense placement, changing sizes, and a wide variety of challenging relationships, there is still much room for improvement in the accuracy of current 3D object detection methods.

\vspace{1mm}
\noindent {\bf 3D Vision and Language:} Compared to the significant progress made in joint inference of language and images, connecting 3D vision to natural language is a relatively new research topic. A recent dataset~\cite{3DVL_SUNspot} combines RGB-D images with language expression to explore the potential gains from adding depth channel beyond a single RGB image. Kong et al.~\cite{3DVL_text2image} exploit language description of single-view RGB-D images of scenes to align the nouns/pronouns with the referred visual objects. Chen et al.~\cite{3DVL_Text2shape} presented conditional generation of 3D models from text, which could be useful in augmented reality applications. Achlioptas et al.~\cite{ReferIt3D} introduced a new large-scale dataset and task that identifies the specific object instances from the same category with known 3D bounding boxes. Unlike these methods, we focus on 
3D object grounding task extended from ScanRefer~\cite{3DVL_Scanrefer} 
%
by capturing the rich structure in the free-form description, extracting the object-object and object-scene co-occurrence relationships in the 3D scene and encoding global contexts of phrases and object proposals by a nodes matching strategy.

\begin{figure*}[t!] 
	\center{\includegraphics[width=1\textwidth]{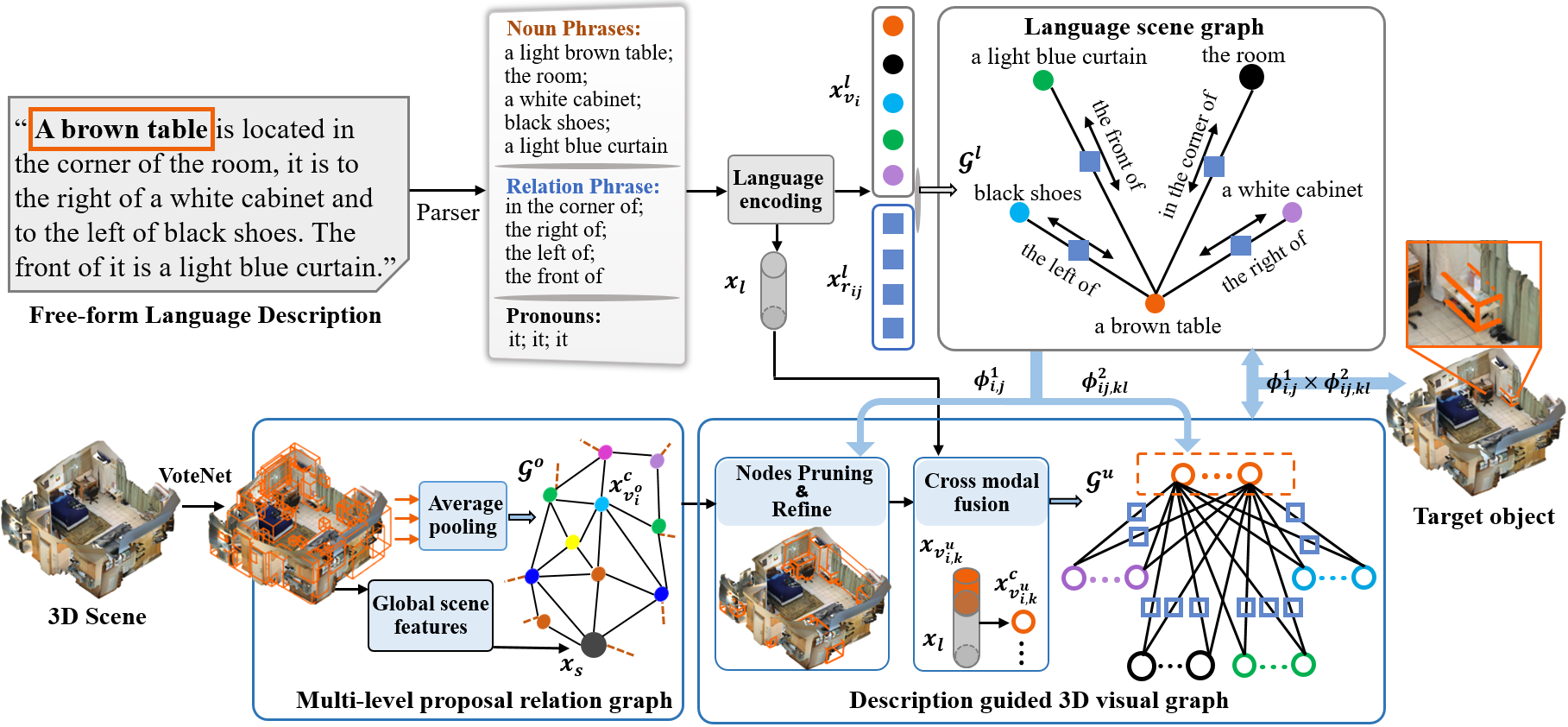}}
	\caption{\label{fig:Framewor} An overview of our proposed network. There are three modules in our method, the language scene graph $\mathcal{G}^l$ incorporates the rich structure and language context; the multi-context proposal relation graph leverages two occurrence relationships (object-object and object-scene) to strength the visual features of the initial proposals set; the description guided 3D visual graph $\mathcal{G}^u$ is defined on the pruned and refined proposals which is under the guidance of $\mathcal{G}^l$, then the nodes of $\mathcal{G}^u$ are adaptively matched with the nodes of $\mathcal{G}^l$, and then we fuse this with the matching score in proposals pruning for the final 3D object grounding. }
	\vspace{-2mm}	
\end{figure*}

\vspace{-2mm}
\section{Proposed Approach}
\vspace{-2mm}
Figure~\ref{fig:Framewor} shows an overview of our free-form description guided 3D object grounding method. It includes a language scene graph module, a multi-context proposal relation graph module and a description guided 3D visual graph module.

\vspace{-1mm}
\subsection{Language Scene Graph Module}
\vspace{-2mm}
Unlike single sentences used in most image-language tasks~\cite{Propagating,Phrase-GD,PhraseCut,Relat-Embed}, we adopt free-form language descriptions composed of several sentences in our work, where additional challenges such as long-distance phrase correlations emerge. Specifically, we first build a language scene graph $\mathcal{G}^l$ from the free-form 3D scene description $L$ to capture the rich structure and relationships between the phrases.

\vspace{1mm}
\noindent {\bf (1) Parsing free-form descriptions:}
As the language descriptions tend to describe not only the properties of the referred 3D object, but also its relationships with nearby 3D objects or the 3D scene, we parse the complex free-form language description into three sub-components by adopting an off-the-shelf Language Scene Graph Parser~\cite{Parser}. We refine the Language Scene Graph Parser~\cite{Parser} for our free-form language description guided 3D visual grounding task by the rule-based post-processing. 

\noindent \textbf{Noun phrases:} We first build a noun library using ground truth 3D object classes of interest in the training dataset, and then add synonyms and plural to the noun library. We extract nouns from the free-form description, and match them to the 3D object classes. We connect nouns not only to the 3D object class, but also to its synonyms and plural forms. To get 3D object attributes for a noun, such as color, shape and size, we search all attribute words that modify a noun of interest in the Scene Graph Parser~\cite{Parser} dependency information~\cite{3DVL_text2image}. We generate noun phrases $\left\{v_i^l\right\}$ by combing the attribute words and the noun, e.g. for ``\textit{The table is light brown}'' we have a noun phrase ``\textit{light brown table}''.  

\noindent \textbf{Pronouns:} We consider free-form descriptions composed of several sentences which are not independent but typically refer to the same 3D object multiple times. Therefore, we are faced with the long-distance coreference problem. For example, in ``\textit{A light brown table is located in the corner of the room, it is to the right of a white cabinet}'', both ``\textit{table}'' and ``\textit{it}'' refer to the same 3D object and thus form a coreference. According to the statistical analysis of the free-form descriptions in dataset, we found that the pronouns refer to the referred 3D object in most cases. To address this, we extract part of speech of all words in a description $L$ using the Language Scene Graph Parser~\cite{Parser}, and replace pronouns with noun phrases of the referred 3D object in the descriptions, so that free-form descriptions which contain multiple sentences can deal with the long-distance correlations.

\noindent \textbf{Relation phrases:} To accurately parse the relation phrases $\{r_{ij}^l\}$ connecting noun phrases $v_i^l$ and $v_j^l$ in a free-form description, we build a relation phrase library from the training dataset. When a relation word can not be directly parsed, we look up all related words in the library and match it with the most similar word, and then expand it according to the standard form.

\vspace{1mm}
\noindent {\bf (2) Language Scene Graph Construction:}
Given a free-form description $L$, we first use the parsed sub-components to construct an initial language scene graph, where each node and edge of the graph correspond to an object and a relationship between this object and another object mentioned in $L$ respectively. We define the language scene graph as a directed graph $\mathcal{G}^l=\{\mathcal{V}^l,\mathcal{R}^l\}$, where $\mathcal{V}^l=\left\{v_i^l\right\}_{i=1}^{I}$ and $\mathcal{R}^l=\left\{r_{ij}^l\right\}$ are nodes and edges set. As each object is represented as an noun phrase with a set of attributes, we first encode each word in description $L$ into a sequence of word embedding $\{h_t\}_{t=1}^T$ using GLoVe~\cite{Glove}, where $T$ is the number of words in description $L$. We then encode the complete noun phrase by taking average pooling using the last hidden state of a Bi-directional GRU~\cite{BIGRU}. Finally, we obtain noun phrase embedding $x_{v_i^l}$ for the corresponding node $v_i^l$ in graph $\mathcal{G}^l$. Similar to the noun phrase, we also compute a vector representation $x_{r_{ij}^l}$ for edge $r_{ij}^l$ in graph $\mathcal{G}^l$.

Furthermore, we use another Bi-directional GRU~\cite{BIGRU} to encode the complete description $L$ and obtain the description presentation $x_l$, which provides the global context information missing in the encoding of individual noun phrases. In addition, fused with 3D point cloud features, this description presentation $x_l$ is then used to construct the 3D visual graph.

\vspace{1mm}
\noindent {\bf (3) Phrase feature refinement}
Inspired by the message propagation operators proposed in ~\cite{Corss-modal}, we first learn the context-aware features for all edges with its connected nodes in graph $\mathcal{G}^l$, where all nodes are associated with embedded features of noun phrases. We aggregate messages from nodes to update the relation features of their corresponding edge:
\begin{equation} 
x_{r_{ij}^l}^c=x_{r_{ij}^l}+F_r^l([x_{v_i^l};x_{v_j^l}; x_{r_{ij}^l}] , \label{RelationP_fea_refine} 
\end{equation}
where $x_{r_{ij}^l}^c\in \mathbb{R}^{D_1}$ is the refined relation feature, and $F_r^l$ is a multilayer network with fully connected layers.

Then, we update each node $v_i^l$ in graph $\mathcal{G}^l$ by aggregating messages from connected all nodes $\mathcal{N}(i)$ and edges to it via self-attention mechanism~\cite{GAT}: 
\begin{equation} 
x_{v_i^l}^c=x_{v_i^l}+\sum_{j\in \mathcal{N}(i)}w_{v_{ij}}F_v^l([x_{v_j^l}; x_{r_{ij}^l}^c]) , \label{NounP_fea_refine} 
\end{equation}
where $x_{v_i^l}^c$ is the refined noun phrase feature, $F_v^l$ is a multilayer network with fully connected layers, and $w_{v_{ij}^l}$ is an attention weight between node $v_i^l$ and $v_j^l$, which is defined as follows:
\begin{equation} 
w_{v_{ij}^l}=\mathop{\rm softmax}\limits_{j\in \mathcal{N}(i)}(F_v^l([x_{v_j}; x_{r_{ij}}^c])^TF_v^l([x_{v_j}; x_{r_{ij}}^c])), \label{NounP_fea_weight} 
\end{equation}
where $\rm softmax$ computes the normalized attention values.

\vspace{-1mm}
\subsection{Multi-level 3D proposal relation graph module} 
\vspace{-2mm}
As the first visual processing step of our model, we need to use a 3D object detection framework to predict an initial set of 3D object proposals. The quality of the 3D object proposals obtained in this step can severely affect the subsequent 3D object grounding performance. In fact, since the state-of-the-art 3D object detection methods still yield limited performance in real world 3D scenes, generally large misalignment exists between the generated 3D bounding box candidates and the ground-truth 3D objects. This could impede the learning procedure of the following stage. Leveraging two co-occurrence relationships (object-object and object-scene), we introduce the multi-level 3D proposal relation graph to encode the context dependency among the global information, which strengthens the visual features of the initial proposals by incorporating co-occurrence relation cues in the 3D scene.

We adapt the VoteNet~\cite{VoteNet} backbone to process the input point cloud and output a set of point clusters with enriched appearance features $x_a^o\in \mathbb{R}^{K_o\times D_2}$, where $K_o$ is the number of proposals. Next, the 3D object detection module takes in the point clusters and processes them to predict 3D bounding boxes candidates $\mathcal{B}=\{b_i\}_{i=1}^{K_o}$ and its object class for all $K_o$ proposals, where each $b_i\in \mathbb{R}^{D_3}$ denotes 3D object location. Additionally, we use average pooling to compute a global scene feature $x_s\in \mathbb{R}^{D_2}$ depending on the set of point clusters. 
To encode the spatial features for each proposal, we combine two types of features, geometric structure feature $S_b$ represented by the relative parameters of each bounding box and relative spatial location feature $L_b$ represented by the relative center coordinates of each bounding box. To obtain strong representational power that captures co-occurrence relationships between different proposals, both appearance and spatial features are considered: 
\begin{equation} 
x_{pv}^o=F_{vf}^o([x_a^o;F_{p}^o([S_b;L_b])]), \label{Obj_fea} 
\end{equation}
where $x_{pv}^o\in\mathbb{R}^{D_4}$ is the 3D proposal visual feature, $F_{vf}^o$ and $F_{p}^o$ denote MLP layers.

We define the multi-level 3D proposal relation graph as $\mathcal{G}^o=\{\mathcal{V}^o,\mathcal{R}^o\}$, where $\mathcal{V}^o=\{v_i^o\}_{i=1}^{K_o+1}$ is the set of nodes including the initial 3D object proposals and the global scene; $\mathcal{R}^o=\{r_{ij}^o\}_{i,j=1}^{K_r}$ is the set of edges, and $r_{ij}^o$ is the edge from $v_i^o$ to $v_j^o$. Considering that an object typically only interacts with the global scene and objects nearby, the graph $\mathcal{G}^o$ aggregates features from all its neighbour nodes $\mathcal{N}(i)$ and global scene via self-attention mechanism~\cite{GAT} to augments the input visual features,
\begin{equation} 
x_{v_i^o}^c=x_{pv}^o+\sum_{j\in \mathcal{N}(i)} r_{ij}^o F_f (x_{pv}^o) , \label{Obj_fea_argument} 
\end{equation}
where $x_{v_i^l}^o$ is the augmented context-aware 3D object visual feature, $F_f$ denote MLP layers, and $r_{ij}^o$ denotes the self-attention weight~\cite{GAT} between nodes $v_i^o$ and $v_j^o$.

\vspace{-0mm}
\subsection{Description guided 3D visual graph module}
\vspace{-1mm}
Although the visual features of the initial candidates in the previous step have been further refined, there are still hundreds of proposals and a lot of noise, which makes it infeasible to identify the target object by exploring the knowledge of language scene graph $\mathcal{G}^l$. We demonstrate this by introducing a description guided 3D visual graph to capture the global scene context via message propagation for few selected proposals, and to reduce the gap between 3D visual information to language structure.

\vspace{1mm}
\noindent {\bf (1) Nodes pruning and refinement:} 
Most 3D object proposals are unlikely to have relationships with the language scene graph $\mathcal{G}^l$ because the object proposals in large 3D scenes are usually redundant. To model these regularities, we introduce a description guided nodes pruning module which exploit the knowledge of our language scene graph $\mathcal{G}^l$ to efficiently estimate the relatedness of the noun phrase nodes and the 3D object proposal nodes, and refine the parameters of the selected 3D bounding boxes. To achieve this, we compute a matching score $\Phi_{ij}^1$ for each noun phrase and 3D object proposal pair, and a noun phrase guided bounding box offset $\psi_{i,j}\in \mathbb{R}^{D_3}$ depending on each noun phrase feature $x_{v_i^l}^c$ and 3D object visual feature $x_{v_i^o}^c$,
\begin{equation} 
\Phi_{i,j}^1=H_{p}^n(x_{v_i^l}^c,x_{v_i^o}^c) , \label{Pruning_probability} 
\end{equation}
\begin{equation} 
\psi_{i,j}=H_{reg}^b(x_{v_i^l}^c,x_{v_i^o}^c)  , \label{Pruning_offset} 
\end{equation}
where $H_{p}^n$ and $H_{reg}^b$ are learning steps which transform the input features as in~\cite{Natural}.

After obtaining the matching scores $\Phi_{ij}^1$ for all noun phrase and 3D object proposal pairs, we sort the scores to choose the top $K$ proposals for each noun phrase and obtain their new 3D bounding boxes $\mathcal{B}'=\{b_i\}_{i=1}^{K}$ according to the regression offset $\psi_{i,j}$. We represent the refined proposal set of each noun phrase $v_i^l$ as $\mathcal{V}_i^u=\{v_{i,k}^u\}_{k=1}^K$. We then apply 3D Non-Maximal Suppression (NMS) to filter out proposals that have significant overlap with others. We obtain the selected 3D object proposal nodes which are much sparser and more accurate than the initial set of proposals.


\vspace{1mm}
\noindent {\bf (2) 3D visual graph:}
Instead of using noisy dense 3D object proposals of ScanRefer~\cite{3DVL_Scanrefer}, we introduce a 3D visual graph $\mathcal{G}^u=\{\mathcal{V}^u,\mathcal{R}^u\}$ on top of the refined set of 3D object proposals. Similar to Eq.~(\ref{Obj_fea}), we extract an new visual feature $x_{v_{i,k}^u}$ for each proposal by fusing its appearance feature and spatial feature. We then concatenate the new visual feature $x_{v_{i,k}^u}$ with the holistic feature $x_l$ of the description as the final node feature $x_{v_{i,k}^u}^c$ of 3D visual graph. Moreover, we compute the edge feature $x_{v_{ij,kl}^u}$ of two proposal $v_{i,k}^u$ and $v_{j,l}^u$ using the minimum box region that covers them. We adopt a set of message propagation operators to generate context aware representations for all the nodes and edges in the graph $\mathcal{G}^u$.
Similar to Eq.~(\ref{RelationP_fea_refine}), Eq.~(\ref{NounP_fea_refine}), and Eq.~(\ref{NounP_fea_weight}), we obtain the updated relation features $x_{v_{ij,kl}^u}^c$, context-aware object feature $x_{v_{i,k}^u}^c$ using the respective types of message propagation operators.

\vspace{-1mm}
\section{Prediction model}
\vspace{-2mm}
Given the language scene graph and 3D visual graph, we formulate the 3D grounding as a node matching problem between the 3D visual graph $\mathcal{G}^u$ and the language scene graph $\mathcal{G}^l$. We introduce a matching score $\Phi_{i,j}^2$ for each noun phrase $v_i^l$ and proposal $v_j^u$ pair,
\begin{equation} 
\Phi_{i,j}^2=H_{p}^u(x_{v_i^l}^c,{v_i^o}^c) , \label{predict_probability} 
\end{equation}
where $H_{p}^u$ is a two layer MLP network.
We then fuse this with the score $\Phi_{i,j}^1$ used in object pruning to generate the node matching score $\Phi_{i,j}=\Phi_{i,j}^1\times \Phi_{i,j}^2$. Finally, the box with the highest score is selected from the $K$ proposal boxes as the target 3D bounding box of the noun phrase. Here, only the target 3D bounding box corresponding to the subject is selected.

\textbf{Loss function.} The final loss is a linear combination of the vote loss $\mathcal{L}_{vt}$, abjectness of loss $\mathcal{L}_{obj}$, bounding box loss $\mathcal{L}_{b}$, semantic classification loss $\mathcal{L}_{sm}$, description classification loss $\mathcal{L}_{cls}$ and reference loss $\mathcal{L}_{rf}$:
\begin{equation} 
\mathcal{L}= \lambda_1 \mathcal{L}_{vt}+\lambda_2 \mathcal{L}_{obj}+\lambda_3 \mathcal{L}_{b}+\lambda_4 \mathcal{L}_{sm}+\lambda_5 \mathcal{L}_{cls}+\lambda_6 \mathcal{L}_{rf}, \label{Loss_final} 
\end{equation}
where $\lambda_1$, $\lambda_2$, $\lambda_3$, $\lambda_4$, $\lambda_5$ and $\lambda_6$ are the weights for the individual loss terms. Specially, $\mathcal{L}_{vt}$ supervises the vote regression step defined in~\cite{VoteNet}, $\mathcal{L}_{obj}$ represents whether the point clusters obtained by voting and aggregation belong to a certain instance object, $\mathcal{L}_{b}$ supervises the box center regression, classifying
the box size classification and the box size regression processes respectively, $\mathcal{L}_{sm}$ supervises the semantic classification process for the $N$ ScanNet dataset classes, $\mathcal{L}_{cls}$ is applied for an object classification based on the input description, $\mathcal{L}_{rf}$ supervises the similarity scores and offset respectively (details in supplementary material).

\vspace{-1mm}
\section{Experiments}
\vspace{-2mm}
We introduce two experimental 3D object grounding datasets and the implementation details of our method. We perform detailed analysis of our method to demonstrate the efficacy of the proposed modules and compare the performance with state-of-the-art.

\begin{table}[t]  
\footnotesize
	\centering
	\begin{tabular}{l|ccc}
		\toprule
		& \multicolumn{1}{c}{Unique} & \multicolumn{1}{c}{Multiple} & \multicolumn{1}{c}{Overall} \\
		Methods             & Acc@0.5      & Acc@0.5      & Acc@0.5     \\
		\midrule
		One-stage~\cite{One-stage}      & 22.82 & 6.49 & 9.04  \\
		VoteNet~\cite{VoteNet} + rand.  & 19.35 & 2.81 & 5.28  \\
		Ours (GT boxes) 	& 75.40  & 30.20 & 43.16 \\
		\midrule
		\midrule
		Ours (xyz) 			 & 64.04  & 24.13 & 32.47 \\
		Ours (xyz+rgb)		 & 66.87 & 25.00  & 33.55 \\
		Ours (xyz+rgb+nor.)  & 67.94  & 25.70  & 34.01  \\		
		\midrule         
		\midrule         
		Ours (w/o LSG) 		& 66.92  & 23.15 & 32.87 \\
		Ours (w/o MLPG)		& 67.10 & 24.98  & 33.14 \\
		Ours (w/o LGVG)     & 65.34 & 23.75 & 31.90  \\
		\midrule         
	\end{tabular}
	\caption{\label{Ablation} Ablation studies on ScanRefer~\cite{3DVL_Scanrefer} validation set. We measure the percentage of predictions whose IoU with the ground truth boxes is greater than 0.5. We also report scores on ``unique'' (single object of class in scene) and ``multiple'' subsets.}
	\vspace{-3mm}
\end{table}

\begin{table*}[] 
	\centering
	\footnotesize
	\begin{tabular}{l|c|cccccc}
		\toprule
		& & \multicolumn{2}{c}{Unique} & \multicolumn{2}{c}{Multiple} & \multicolumn{2}{c}{Overall} \\
		Methods& Input  & Acc@0.25     & Acc@0.5     & Acc@0.25      & Acc@0.5      & Acc@0.25      & Acc@0.5     \\
		\midrule		
		One-stage~\cite{One-stage} 	     &2D image             & 29.32 & 22.82  & 18.72 & 6.49 & 20.38 & 9.04 \\
		IntanceRefer~\cite{InstanceRefer}    &xyz+rgb+normals  & \underline{77.13} & \underline{66.40} & 28.83 & \underline{22.92}  & 38.20 & \underline{31.35} \\
		ScanRefer~\cite{3DVL_Scanrefer}	 &xyz+rgb+normals      & 67.64 & 46.19 & \underline{32.06} & 21.26  & \underline{38.97} & 26.10 \\
		Ours        &xyz+rgb+normals                           & \textbf{78.80} & \textbf{67.94} & \textbf{35.19} & \textbf{25.70} & \textbf{41.33}  & \textbf{34.01}  \\
		\midrule         
	\end{tabular}
	\caption{\label{Compar_ScanRefer} Comparison with state-of-the-art on ScanRefer~\cite{3DVL_Scanrefer} dataset. Best performance is marked in bold and the second best is underlined. }
		\vspace{-4mm}
\end{table*}

\begin{table}[] 
	\centering
	\footnotesize
	\begin{tabular}{l|ccccc}
		\toprule
		Method & Overall & Easy & Hard & VD & VI \\
		\midrule
		ScanRefer~\cite{3DVL_Scanrefer}     & 34.2 & 41.0 & 23.5 & 29.9 & 35.4  \\
		IntanceRefer~\cite{InstanceRefer}   & \underline{38.8} & \underline{46.0} & \underline{31.8} & \underline{34.5} & \underline{41.9}  \\
		ReferIt3D~\cite{InstanceRefer}      & 35.6 & 43.6 & 27.9 & 32.5 & 37.1  \\		
		Ours                                & \textbf{41.7} & \textbf{48.2} & \textbf{35.0} & \textbf{37.1} & \textbf{44.7}  \\
		\midrule 
	\end{tabular}
	\caption{\label{Compar_Nr3D} Comparison with state-of-the-art on Nr3D~\cite{ReferIt3D} dataset. Here ``easy'' and ``hard'' are determined by whether there are more than two instances of the same object class in the scene. VD/VI mean view-dependent/view-independent i.e. whether the language description depends on the camera view or not.}
	\vspace{-6mm}
\end{table}

\subsection{Datasets}
\vspace{-2mm}
Experiments were performed on ScanRefer~\cite{3DVL_Scanrefer} and Nr3D~\cite{ReferIt3D} datasets, which are built on top of ScanNet~\cite{ScanNet}.

\noindent \textbf{ScanRefer} contains $51,583$ descriptions of $11,046$ 3D objects for 800 ScanNet~\cite{ScanNet} real world scenes. It is the first large-scale dataset to induct 3D object grounding in point clouds via complex and diverse natural language descriptions. There are an average of $13.81$ objects, $64.48$ descriptions per scene, and $4.67$ descriptions per object. Following the official ScanRefer~\cite{3DVL_Scanrefer} splits, we use the training set of $36,665$ descriptions and validation set of $9,508$ descriptions. Each description is annotated by a ground truth 3D bounding box presented by center coordinates, orientations and dimensions in an indoor scene. 
 
\noindent \textbf{Nr3D} is a real-world 3D scene dataset with extensive free-form natural language descriptions. It contains $41,503$ human utterances collected by deploying an online reference game in AMT. Following the official Nr3D~\cite{ReferIt3D} split, we use the training set of $29,500$ descriptions and validation set of $7,650$ descriptions. Each description is annotated by a ground truth 3D bounding box presented by center coordinates, orientations and dimensions in an indoor scene.  

To augment the training data, we follow the operations in~\cite{VoteNet}, randomly flip each scene in both horizontal direction, randomly rotate the scene points around the Z-axis by an angle selected uniformly between $[-30^\circ; 30^\circ]$ and globally scale the scene points between $[0.9; 1.1]$. Results and analysis are reported for the validation split.

\vspace{-1mm}
\subsection{Implementation details}
\vspace{-2mm}

We generate an initial set of $K_o=256$ 3D object proposals using VoteNet~\cite{VoteNet}. In graph $\mathcal{G}^l$ module, the refined relation feature dimension $D_1$ is set to 128. In graph $\mathcal{G}^o$ module, the dimensions of the appearance feature $D_2$ is set to 256; since each 3D bounding box is parameterized by the box center, the box size and semantic classes, the parameter dimension $D_3$ is set to 24; the 3D proposal visual feature dimension is set to 128. In graph $\mathcal{G}^u$ module, the selected proposals number $K$ is set to 20. For model training, we use ADAM optimizer with initial learning rate $1e-3$. We train 30 epoches with batch size 32 and decay the learning rate 10 times after 5, 15 and 25 epoches. The loss weights of regression terms $\lambda_1$ and $\lambda_3$ are set to 1 while $\lambda_2$ is set to 0.5 and $\lambda_4=\lambda_5=\lambda_6$ are set to 0.1.

\vspace{-1mm}
\subsection{Ablation studies}
\vspace{-2mm}

We select the ScanRefer~\cite{3DVL_Scanrefer} dataset, as it contains complex and multi-sentences descriptions making it more challenging, to conduct three groups of ablation studies

\noindent \textbf{Task complexity:} The top part of Table~\ref{Ablation} shows the effect of using a 2D object grounding method~\cite{One-stage} where 
the 2D proposal with the highest confidence score is projected to 3D using the recorded camera parameters for that view. Grounding in 2D images results in inaccurate 3D bounding boxes as it suffers from the limited view of the 3D scene. Hence, it is necessary to design a method for direct object grounding in a 3D scene. Next, we randomly selected one of the proposals from the VoteNet~\cite{VoteNet} backbone that matches the ground truth semantic class label and found that it is insufficient to identify the referred 3D object from only the semantic label. Furthermore, we investigate the case where our model uses ground truth 3D bounding boxes and observe a huge improvement of $24.97\%$ on overall Acc$@0.5$. This confirms that current 3D object detection backbones have a large room for improvement and that our description guided 3D visual graph module is necessary to refine the initial 3D bounding box candidates.


\noindent \textbf{Different inputs:} We conduct an ablation study on our model to examine what components and point cloud features contribute to the performance. Results are reported in the middle part of Table~\ref{Ablation}. The performance of our model improves when $rgb$ information is added to the location $xyz$ features as input. The performance further improves when normals ($nor.$) are also included from the ScanNet~\cite{ScanNet} meshes. Color information and additional geometric information can enhance the performance of our network, because they are usually included in noun phrases as attributes in the language description and as part of the features of the language scene graph nodes. If the input point cloud contains both color information and additional geometric information, the expressive power of the 3D visual scene graph becomes stronger, and higher similarity matching scores will be obtained during the two matching processes, thereby improving the final prediction accuracy.  

\noindent \textbf{Units effectiveness:} We conducted an ablation study on the effect of the proposed modules in our model. Results are reported in the bottom part of Table~\ref{Ablation}. We set the baseline when our complete framework uses spatial coordinates, color and normal information as input (i.e. $xyz+rgb+nor.$). When the language scene graph module (LGS) or the multi-level proposal graph module (MLPG) is not used, there is a drop in performance over the baseline. This is mainly because the context-aware  phrases presentation in language description and the mutual occurrence relationships has the ability to better handle complex scenes. There is a higher drop in performance over the baseline when the description guided 3D visual graph module (LGVG) is not used. 
The reason is that, with complex and diverse descriptions, the relationships between 3D object proposals and the phrase can be ambiguous without LGVG. 


\vspace{-1mm}
\subsection{Quantitative comparisons}
\vspace{-2mm}
We first compare our method with the state-of-the-art approaches including One-stage~\cite{One-stage}, IntanceRefer~\cite{InstanceRefer} and ScanRefer~\cite{3DVL_Scanrefer} on ScanRefer~\cite{3DVL_Scanrefer} dataset in Table~\ref{Compar_ScanRefer}. Particularly, One-stage~\cite{One-stage} is a 2D image based method, which predicts the referred 2D object in each frame of the scan video. Then they select the 2D bounding box with the highest probability value and project it to 3D space using the camera parameters of that frame. Note that IntanceRefer~\cite{InstanceRefer}, ScanRefer~\cite{3DVL_Scanrefer} and our method all use coordinates, RGB information and normals of point clouds as input. We report the precision using the percentage of predictions whose IoU with the ground truth boxes is greater than 0.25 and 0.5 as the evaluation metric. As summarized in Table~\ref{Compar_ScanRefer}, our approach outperforms the prior methods by large margins. 2D based method One-stage~\cite{One-stage} cannot achieve satisfactory results since it is limited by the view of a single frame. It is worth noting that our method achieves a $\sim1.5\%$ improvement of Acc$@$0.5 for scenes with a single object of its class, and achieves a remarkable $\sim2.8\%$ improvement of Acc$@$0.5 for more complex scenes containing multiple classes of objects. Furthermore, we observe that our proposed method performs better by a larger margin in ``Multiple'' than ``Unique'' cases. This supports our claim that our proposed pipeline comprising three novel modules can better handle complex interactions between 3D objects and the cross-modal communications from free-form descriptions to point clouds.

\begin{figure*}[t!] 
	\center{\includegraphics[width=0.95\textwidth]{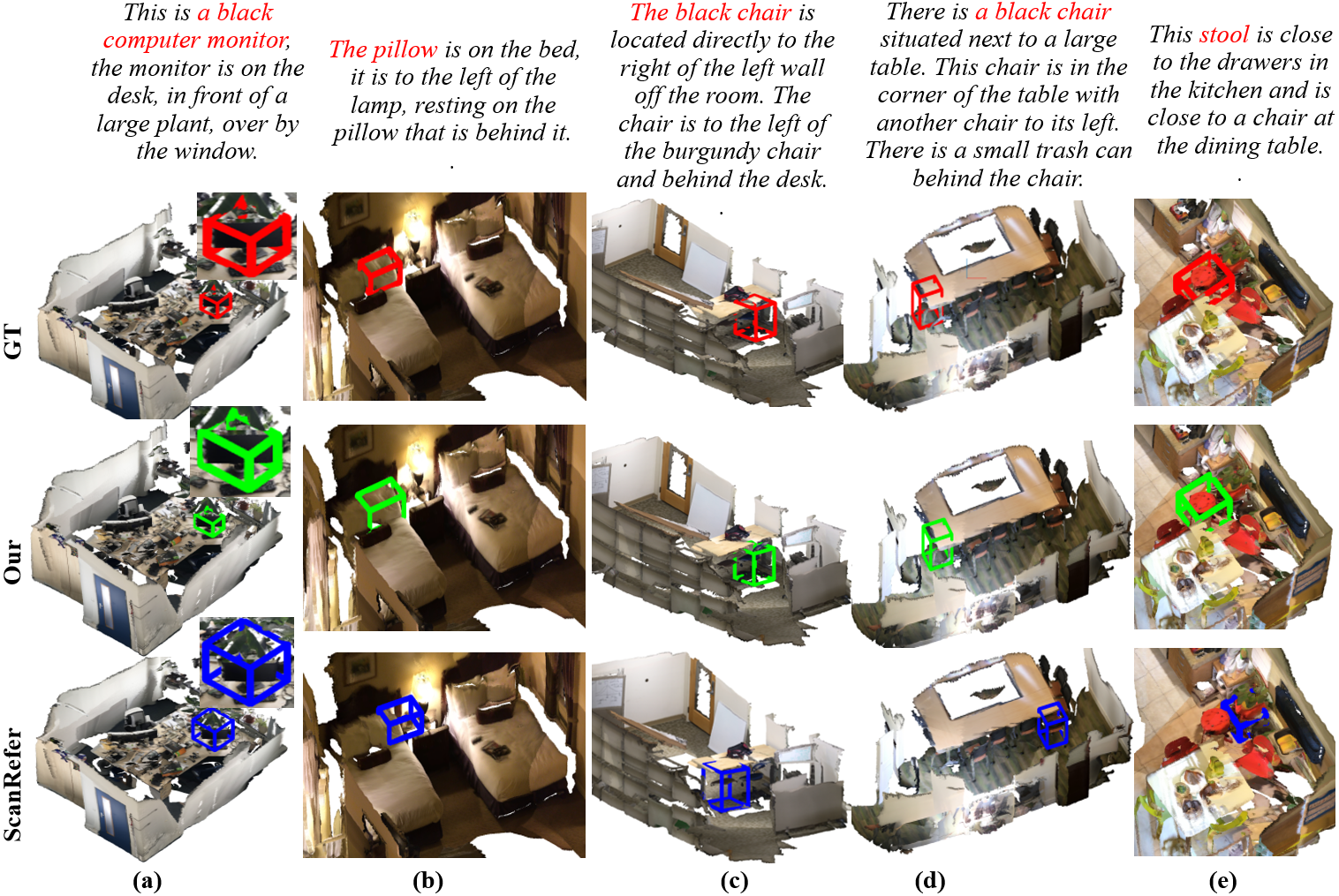}}
	\vspace{-1mm}
	\caption{\label{fig:Detection_vis} Results of ScanRefer~\cite{3DVL_Scanrefer} method and our method on ScanRefer~\cite{3DVL_Scanrefer} dataset (columns 1-4) and Nr3D~\cite{ReferIt3D} dataset (last column). 
	}
	\vspace{-4mm}
\end{figure*}

\begin{figure}[] 
	\center{\includegraphics[width=0.43\textwidth]{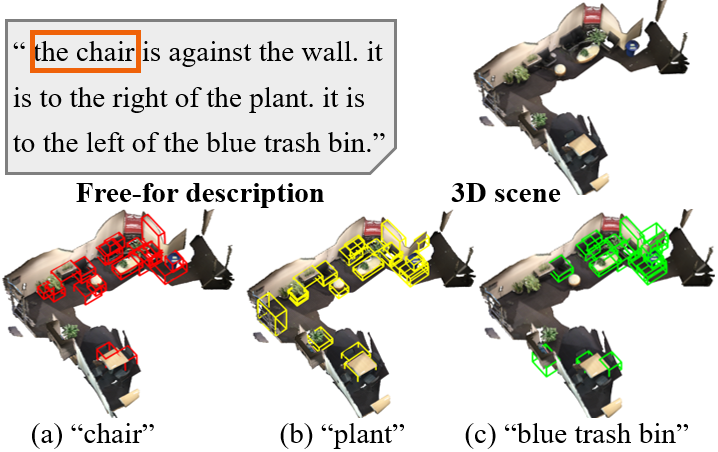}}
	\caption{\label{fig:Model_vis} Results of the most relative 3D bounding boxes for each noun phrase in description guided 3D visual graph module.}
	\vspace{-5mm}	
\end{figure}

Table~\ref{Compar_Nr3D} compares our method with state-of-the-art methods ScanRefer~\cite{3DVL_Scanrefer}, IntanceRefer~\cite{InstanceRefer} and ReferIt3D~\cite{ReferIt3D} on the Nr3D~\cite{ReferIt3D} dataset. ReferIt3D~\cite{ReferIt3D} assumes that ground truth 3D bounding boxes for each 3D scene are already given in the input, and the task is just to select which 3D bounding box is the referred 3D object. As shown in Table~\ref{Compar_Nr3D}, our method achieves the best performance, which are $2.9\%$ higher than IntanceRefer~\cite{InstanceRefer}, $6.1\%$ higher than ReferIt3D~\cite{ReferIt3D}, and $7.5\%$ higher than ScanRefer~\cite{3DVL_Scanrefer}. Also note that our method has the highest gain in accuracy for `Hard' cases which again supports our claim that our proposed framework comprising the three novel modules can effectively handle complex scenes and free-form descriptions.
Furthermore, our graph formulation of 3D visual information is more effective than ReferIt3D~\cite{ReferIt3D} in modeling the free-form description and geometric context.

\vspace{-2mm}
\subsection{Visualization}
\vspace{-2mm}
\noindent \textbf{Qualitative Visualization results:} Figure~\ref{fig:Detection_vis} shows 5 qualitative visual grounding results produced by the ScanRefer~\cite{3DVL_Scanrefer} method and our method on the ScanRefer dataset (first four columns) and the Nr3D dataset (last column). The successful detection of objects in the first two columns show that our multi-level 3D proposal graph module and description guided 3D visual module can handle the spatial relations to regress more accurate 3D bounding boxes than ScanRefer. The failure cases in the last three columns show that ScanRefer is unable to distinguish ambiguous objects in 3D scenes with complex and diverse descriptions. The performance of ScanRefer is limited since it fuses holistic language representations with visual features and ignores the relationships between proposals and phrase.   

\noindent \textbf{Model visualization:} We visualize the results of the top $K=20$ 3D proposals selected by the relevant noun phrase in our description guided 3D visual graph module, as shown in Figure.~\ref{fig:Model_vis} second row. It is evident that each noun phrase node in language scene graph matches all instances of the same category and all objects that have a strong relationship with it in the scene, which prunes and refines the initial redundant proposals set to boost the performance of the subsequent steps.

\vspace{-2mm}
\section{Conclusion}
\vspace{-2mm}
We proposed a free-form description guided 3D visual graph network for 3D object grounding in point clouds. Our method achieves accurate detection via capturing the intra-modal and cross modal relationships between the natural language descriptions and the 3D scenes. The complex free-form description is first parsed and then a language scene graph is constructed to compute a context-aware phrases presentation through message propagation.
A multi-level 3D relation graph was introduced to leverage two co-occurrence relationships (object-object and object-scene) and strengthen the visual features of the initial proposals. A 3D visual graph was constructed over the refined proposals to encode global contexts of phrases and proposals under the guidance of language scene graph. Experiments on two challenging benchmark datasets show that our method quantitatively and qualitatively outperforms existing state-of-the-art in 3D object grounding.

{\small
\bibliographystyle{ieee_fullname}
\bibliography{egbib}

\begin{thebibliography}{10}\itemsep=-1pt

\bibitem{Parser}
https://github.com/vacancy/scenegraphparser.

\bibitem{ReferIt3D}
Panos Achlioptas, Ahmed Abdelreheem, Fei Xia, Mohamed Elhoseiny, and Leonidas
  Guibas.
\newblock Referit3d: Neural listeners for fine-grained 3d object identification
  in real-world scenes.
\newblock In {\em European Conference on Computer Vision}, pages 422--440,
  2020.

\bibitem{3DVL_Scanrefer}
Dave~Zhenyu Chen, Angel~X Chang, and Matthias Nie{\ss}ner.
\newblock Scanrefer: 3d object localization in rgb-d scans using natural
  language.
\newblock {\em 16th European Conference on Computer Vision (ECCV)}, 2020.

\bibitem{Hierarchical}
Jintai Chen, Biwen Lei, Qingyu Song, Haochao Ying, Danny~Z Chen, and Jian Wu.
\newblock A hierarchical graph network for 3d object detection on point clouds.
\newblock In {\em Proceedings of the IEEE/CVF Conference on Computer Vision and
  Pattern Recognition}, pages 392--401, 2020.

\bibitem{3DVL_Text2shape}
Kevin Chen, Christopher~B Choy, Manolis Savva, Angel~X Chang, Thomas
  Funkhouser, and Silvio Savarese.
\newblock Text2shape: Generating shapes from natural language by learning joint
  embeddings.
\newblock In {\em Asian Conference on Computer Vision}, pages 100--116.
  Springer, 2018.

\bibitem{Multi-view}
Xiaozhi Chen, Huimin Ma, Ji Wan, Bo Li, and Tian Xia.
\newblock Multi-view 3d object detection network for autonomous driving.
\newblock In {\em Proceedings of the IEEE conference on Computer Vision and
  Pattern Recognition}, pages 1907--1915, 2017.

\bibitem{RealT-GD}
Xinpeng Chen, Lin Ma, Jingyuan Chen, Zequn Jie, Wei Liu, and Jiebo Luo.
\newblock Real-time referring expression comprehension by single-stage
  grounding network.
\newblock {\em arXiv preprint arXiv:1812.03426}, 2018.

\bibitem{Cops-Ref-data}
Zhenfang Chen, Peng Wang, Lin Ma, Kwan-Yee~K Wong, and Qi Wu.
\newblock Cops-ref: A new dataset and task on compositional referring
  expression comprehension.
\newblock In {\em Proceedings of the IEEE/CVF Conference on Computer Vision and
  Pattern Recognition}, pages 10086--10095, 2020.

\bibitem{BIGRU}
Junyoung Chung, Caglar Gulcehre, Kyung~Hyun Cho, and Yoshua Bengio.
\newblock Empirical evaluation of gated recurrent neural networks on sequence
  modeling, december 2014.
\newblock {\em arXiv preprint arXiv:1412.3555}, 2019.

\bibitem{ScanNet}
Angela Dai, Angel~X Chang, Manolis Savva, Maciej Halber, Thomas Funkhouser, and
  Matthias Nie{\ss}ner.
\newblock Scannet: Richly-annotated 3d reconstructions of indoor scenes.
\newblock In {\em Proceedings of the IEEE Conference on Computer Vision and
  Pattern Recognition}, pages 5828--5839, 2017.

\bibitem{Sequence-GD}
Pelin Dogan, Leonid Sigal, and Markus Gross.
\newblock Neural sequential phrase grounding (seqground).
\newblock In {\em Proceedings of the IEEE/CVF Conference on Computer Vision and
  Pattern Recognition}, pages 4175--4184, 2019.

\bibitem{RGN_3D}
Mingtao Feng, Syed~Zulqarnain Gilani, Yaonan Wang, Liang Zhang, and Ajmal Mian.
\newblock Relation graph network for 3d object detection in point clouds.
\newblock {\em IEEE Transactions on Image Processing}, 30:92--107, 2020.

\bibitem{Survey_guo}
Yulan Guo, Hanyun Wang, Qingyong Hu, Hao Liu, Li Liu, and Mohammed Bennamoun.
\newblock Deep learning for 3d point clouds: A survey.
\newblock {\em IEEE transactions on pattern analysis and machine intelligence},
  2020.

\bibitem{Referitgame}
Sahar Kazemzadeh, Vicente Ordonez, Mark Matten, and Tamara Berg.
\newblock Referitgame: Referring to objects in photographs of natural scenes.
\newblock In {\em Proceedings of the 2014 conference on empirical methods in
  natural language processing (EMNLP)}, pages 787--798, 2014.

\bibitem{3DVL_text2image}
Chen Kong, Dahua Lin, Mohit Bansal, Raquel Urtasun, and Sanja Fidler.
\newblock What are you talking about? text-to-image coreference.
\newblock In {\em Proceedings of the IEEE conference on computer vision and
  pattern recognition}, pages 3558--3565, 2014.

\bibitem{Pointpillars}
Alex~H Lang, Sourabh Vora, Holger Caesar, Lubing Zhou, Jiong Yang, and Oscar
  Beijbom.
\newblock Pointpillars: Fast encoders for object detection from point clouds.
\newblock In {\em Proceedings of the IEEE/CVF Conference on Computer Vision and
  Pattern Recognition}, pages 12697--12705, 2019.

\bibitem{Corss-modal}
Yongfei Liu, Bo Wan, Xiaodan Zhu, and Xuming He.
\newblock Learning cross-modal context graph for visual grounding.
\newblock In {\em Proceedings of the AAAI Conference on Artificial
  Intelligence}, volume~34, pages 11645--11652, 2020.

\bibitem{Seg-GD1}
Gen Luo, Yiyi Zhou, Xiaoshuai Sun, Liujuan Cao, Chenglin Wu, Cheng Deng, and
  Rongrong Ji.
\newblock Multi-task collaborative network for joint referring expression
  comprehension and segmentation.
\newblock In {\em Proceedings of the IEEE/CVF Conference on Computer Vision and
  Pattern Recognition}, pages 10034--10043, 2020.

\bibitem{Generation}
Junhua Mao, Jonathan Huang, Alexander Toshev, Oana Camburu, Alan~L Yuille, and
  Kevin Murphy.
\newblock Generation and comprehension of unambiguous object descriptions.
\newblock In {\em Proceedings of the IEEE conference on computer vision and
  pattern recognition}, pages 11--20, 2016.

\bibitem{3DVL_SUNspot}
Cecilia Mauceri, Martha Palmer, and Christoffer Heckman.
\newblock Sun-spot: an rgb-d dataset with spatial referring expressions.
\newblock In {\em Proceedings of the IEEE/CVF International Conference on
  Computer Vision Workshops}, pages 0--0, 2019.

\bibitem{Natural}
Lili Mou, Rui Men, Ge Li, Yan Xu, Lu Zhang, Rui Yan, and Zhi Jin.
\newblock Natural language inference by tree-based convolution and heuristic
  matching.
\newblock In {\em ACL}, 2016.

\bibitem{Glove}
Jeffrey Pennington, Richard Socher, and Christopher~D Manning.
\newblock Glove: Global vectors for word representation.
\newblock In {\em Proceedings of the 2014 conference on empirical methods in
  natural language processing (EMNLP)}, pages 1532--1543, 2014.

\bibitem{Flickr30k}
Bryan~A Plummer, Liwei Wang, Chris~M Cervantes, Juan~C Caicedo, Julia
  Hockenmaier, and Svetlana Lazebnik.
\newblock Flickr30k entities: Collecting region-to-phrase correspondences for
  richer image-to-sentence models.
\newblock In {\em Proceedings of the IEEE international conference on computer
  vision}, pages 2641--2649, 2015.

\bibitem{VoteNet}
Charles~R Qi, Or Litany, Kaiming He, and Leonidas~J Guibas.
\newblock Deep hough voting for 3d object detection in point clouds.
\newblock In {\em Proceedings of the IEEE/CVF International Conference on
  Computer Vision}, pages 9277--9286, 2019.

\bibitem{PointNet++}
Charles~R Qi, Li Yi, Hao Su, and Leonidas~J Guibas.
\newblock Pointnet++ deep hierarchical feature learning on point sets in a
  metric space.
\newblock In {\em Proceedings of the 31st International Conference on Neural
  Information Processing Systems}, pages 5105--5114, 2017.

\bibitem{GD-review}
Yanyuan Qiao, Chaorui Deng, and Qi Wu.
\newblock Referring expression comprehension: A survey of methods and datasets.
\newblock {\em IEEE Transactions on Multimedia}, 2020.

\bibitem{3Dobject-survey}
Mohammad~Muntasir Rahman, Yanhao Tan, Jian Xue, and Ke Lu.
\newblock Recent advances in 3d object detection in the era of deep neural
  networks: A survey.
\newblock {\em IEEE Transactions on Image Processing}, 29:2947--2962, 2019.

\bibitem{FasterRCNN}
Shaoqing Ren, Kaiming He, Ross~B Girshick, and Jian Sun.
\newblock Faster r-cnn: Towards real-time object detection with region proposal
  networks.
\newblock In {\em NIPS}, 2015.

\bibitem{PointCNN}
Weijing Shi and Raj Rajkumar.
\newblock Point-gnn: Graph neural network for 3d object detection in a point
  cloud.
\newblock In {\em Proceedings of the IEEE/CVF conference on computer vision and
  pattern recognition}, pages 1711--1719, 2020.

\bibitem{GAT}
Petar Veli{\v{c}}kovi{\'{c}}, Guillem Cucurull, Arantxa Casanova, Adriana
  Romero, Pietro Li{\`{o}}, and Yoshua Bengio.
\newblock Graph attention networks.
\newblock {\em International Conference on Learning Representations}, 2018.

\bibitem{Neighbor_Watch}
Peng Wang, Qi Wu, Jiewei Cao, Chunhua Shen, Lianli Gao, and Anton van~den
  Hengel.
\newblock Neighbourhood watch: Referring expression comprehension via
  language-guided graph attention networks.
\newblock In {\em Proceedings of the IEEE/CVF Conference on Computer Vision and
  Pattern Recognition}, pages 1960--1968, 2019.

\bibitem{PhraseCut}
Chenyun Wu, Zhe Lin, Scott Cohen, Trung Bui, and Subhransu Maji.
\newblock Phrasecut: Language-based image segmentation in the wild.
\newblock In {\em Proceedings of the IEEE/CVF Conference on Computer Vision and
  Pattern Recognition}, pages 10216--10225, 2020.

\bibitem{Pixor}
Bin Yang, Wenjie Luo, and Raquel Urtasun.
\newblock Pixor: Real-time 3d object detection from point clouds.
\newblock In {\em Proceedings of the IEEE conference on Computer Vision and
  Pattern Recognition}, pages 7652--7660, 2018.

\bibitem{Single-view}
Guandao Yang, Yin Cui, Serge Belongie, and Bharath Hariharan.
\newblock Learning single-view 3d reconstruction with limited pose supervision.
\newblock In {\em Proceedings of the European Conference on Computer Vision
  (ECCV)}, pages 86--101, 2018.

\bibitem{DynamicGA}
Sibei Yang, Guanbin Li, and Yizhou Yu.
\newblock Dynamic graph attention for referring expression comprehension.
\newblock In {\em Proceedings of the IEEE/CVF International Conference on
  Computer Vision}, pages 4644--4653, 2019.

\bibitem{Propagating}
Sibei Yang, Guanbin Li, and Yizhou Yu.
\newblock Propagating over phrase relations for one-stage visual grounding.
\newblock In {\em European Conference on Computer Vision}, pages 589--605.
  Springer, 2020.

\bibitem{Relat-Embed}
Sibei Yang, Guanbin Li, and Yizhou Yu.
\newblock Relationship-embedded representation learning for grounding referring
  expressions.
\newblock {\em IEEE Transactions on Pattern Analysis and Machine Intelligence},
  2020.

\bibitem{One-stage}
Zhengyuan Yang, Boqing Gong, Liwei Wang, Wenbing Huang, Dong Yu, and Jiebo Luo.
\newblock A fast and accurate one-stage approach to visual grounding.
\newblock In {\em Proceedings of the IEEE/CVF International Conference on
  Computer Vision}, pages 4683--4693, 2019.

\bibitem{Seg-GD2}
Linwei Ye, Mrigank Rochan, Zhi Liu, and Yang Wang.
\newblock Cross-modal self-attention network for referring image segmentation.
\newblock In {\em Proceedings of the IEEE/CVF Conference on Computer Vision and
  Pattern Recognition}, pages 10502--10511, 2019.

\bibitem{Mattnet}
Licheng Yu, Zhe Lin, Xiaohui Shen, Jimei Yang, Xin Lu, Mohit Bansal, and
  Tamara~L Berg.
\newblock Mattnet: Modular attention network for referring expression
  comprehension.
\newblock In {\em Proceedings of the IEEE Conference on Computer Vision and
  Pattern Recognition}, pages 1307--1315, 2018.

\bibitem{Model_context}
Licheng Yu, Patrick Poirson, Shan Yang, Alexander~C Berg, and Tamara~L Berg.
\newblock Modeling context in referring expressions.
\newblock In {\em European Conference on Computer Vision}, pages 69--85.
  Springer, 2016.

\bibitem{InstanceRefer}
Zhihao Yuan, Xu Yan, Yinghong Liao, Ruimao Zhang, Zhen Li, and Shuguang Cui.
\newblock Instancerefer: Cooperative holistic understanding for visual
  grounding on point clouds through instance multi-level contextual referring.
\newblock {\em arXiv preprint arXiv:2103.01128}, 2021.

\bibitem{Voxelnet}
Yin Zhou and Oncel Tuzel.
\newblock Voxelnet: End-to-end learning for point cloud based 3d object
  detection.
\newblock In {\em Proceedings of the IEEE Conference on Computer Vision and
  Pattern Recognition}, pages 4490--4499, 2018.

\bibitem{Phrase-GD}
Haidong Zhu, Arka Sadhu, Zhaoheng Zheng, and Ram Nevatia.
\newblock Utilizing every image object for semi-supervised phrase grounding.
\newblock In {\em Proceedings of the IEEE/CVF Winter Conference on Applications
  of Computer Vision}, pages 2210--2219, 2021.

\end{thebibliography}
}

\end{document}